# Integrating Specialized Classifiers Based on Continuous Time Markov Chain

**Zhizhong Li** and **Dahua Lin**
The Chinese University of Hong Kong
{lz015, dhlin}@ie.cuhk.edu.hk

## Abstract

Specialized classifiers, namely those dedicated to a subset of classes, are often adopted in real-world recognition systems. However, integrating such classifiers is nontrivial. Existing methods, *e.g.* weighted average, usually implicitly assume that all constituents of an ensemble cover the same set of classes. Such methods can produce misleading predictions when used to combine specialized classifiers. This work explores a novel approach. Instead of combining predictions from individual classifiers directly, it first decomposes the predictions into sets of pairwise preferences, treating them as transition channels between classes, and thereon constructs a continuous-time Markov chain, and use the equilibrium distribution of this chain as the final prediction. This way allows us to form a coherent picture over all specialized predictions. On large public datasets, the proposed method obtains considerable improvement compared to mainstream ensemble methods, especially when the classifier coverage is highly unbalanced.

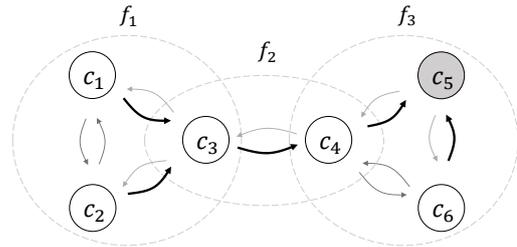

Figure 1: This example illustrates the basic idea of our approach. Here, three specialized classifiers, $f_1$, $f_2$, and $f_3$, respectively cover three subsets of classes $\{c_1, c_2, c_3\}$, $\{c_3, c_4\}$, and $\{c_4, c_5, c_6\}$. Suppose the ground-truth class of a sample is $c_5$, and these classifiers respectively predict $c_3$, $c_4$, $c_5$, based on their restricted scope. According to the preferences conveyed by these predictions, we can derive a set of transition channels, which together constitute a Markov chain. Along this chain, a random walk starting from any class will eventually reach the ground-truth $c_5$. In less ideal cases, where individual classifiers yield conflicting predictions, there can be channels of both directions between a pair of classes. For such cases, we rely on the equilibrium distribution to predict the most likely classes.

## 1 Introduction

Classifier combination is a widely used strategy for improving classification performance. Previous study along this line has resulted in a plethora of methods [Sharkey, 1999; Sewell, 2011; Woźniak *et al.*, 2014]. However, with the recent rise of deep learning [Krizhevsky *et al.*, 2012; Szegedy *et al.*, 2015; Simonyan and Zisserman, 2014; He *et al.*, 2016a], a question regarding the utility of classifier ensemble arises: *Is it still useful to combine classifiers given the strong performance of state-of-the-art deep networks?*

Real-world practice in recent years suggests a positive answer to this question. First, on a number of significant challenges, *e.g.* ImageNet [Russakovsky *et al.*, 2015] and ActivityNet [Fabian Caba Heilbron and Niebles, 2015], top-ranked solutions all rely on ensembles to achieve competitive performance. For example, on ImageNet, an ensemble can often attain over 10% of relative error reduction compared to any individual classifiers, including state-of-the-art networks, *e.g.* Inception-ResNet [Szegedy *et al.*, 2016]. We also observed empirically that combinations of specialized classifiers can often outperform a general classifier, sometimes by a considerable margin. Second, training a deep network usually requires a large dataset and a sheer amount of computational resources. This is not always feasible when factors like time, cost, and data availability are taken into account. Under such circumstances, there can be no other choices but to rely on existing classifiers. It is not uncommon that the classifiers from different sources cover different sets of classes. Therefore, classifier combination remains an effective method to improve performance. It may even be an indispensable part in certain real-world applications.

This paper proposes a novel way to integrate specialized classifiers. Instead of combining their predictions directly, we consider the procedure of reaching the final prediction as a biased random walk among classes. Specifically, we decompose each prediction into a set of *pairwise preferences* and then construct a continuous-time Markov chain by treating such preferences as transition channels, as shown in Figure 1. This way results in a unified picture over all specialized predictions, from which we may draw a more reliable conclusion.

We tested the proposed methods on large datasets. The results show that the presented methods can lead to further performance gains over state-of-the-art deep networks. The predictions derived from these methods are also more accurate than those based on classical combination techniques. It is also noteworthy that the presented formulations are very general and can be directly used in various contexts. For example, one can use them to combine out-of-box classifiers trained for different sets of classes to support a more general application, without re-training.

## 2 Related Work

Extensive study on classifier combination for decades had led to a plethora of methods. See [Sharkey, 1999; Sewell, 2011; Woźniak *et al.*, 2014] for more comprehensive reviews. In general, existing methods fall into two categories: *ensemble methods* and *modular methods*. Specifically, *ensemble methods* combine a set of classifiers that work with the same set of classes; while *modular methods* take advantage of domain-specific knowledges to divide the classification task into sub-tasks, each accomplished by a specialized expert, *e.g.* one expert classifies *flowers*, while another one differentiates between *animals*.

Designs of ensemble methods usually revolve around two desiderata: *accuracy* and *diversity*. To be more specific, on one hand, individual classifiers should yield reasonable predictions; on the other hand, different classifiers should be complementary to each other, such that the combination of their predictions can hopefully result in improved accuracy. Methods that are widely used for constructing diverse ensembles include *Bagging* [Breiman, 1996], *Adaboost* [Freund *et al.*, 1996], and *Random subspace* [Ho, 1998], etc. The way to combine individual predictions is another focus of the research along this line. Combination schemes explored in previous work range from simple aggregation rules such as majority voting, to sophisticated models, such as stacking [Wolpert, 1992], combination based on the Dempster-Shafer theory [E. Mandler, 1988; Rogova, 1994], and error-correcting output coding [Dietterich and Bakiri, 1995].

Modular methods adopt the divide-and-conquer strategy. Particularly, they decompose the original problem into *modules*, and combine their results based on the structures among them. Relations among modules can be either *cooperative*, *i.e.* all modules contribute to the final prediction, or *competitive*, *i.e.* each module has its own realm. In former cases, simple rules such as weighted voting are often used; while for the latter, a particular module needs to be chosen in advance to make the prediction. For example, *Mixture of Experts* [Jacobs *et al.*, 1991; Jordan and Jacobs, 1994] makes this choice via a gating network.

In recent years, classifier combination remains under active research [Karakatič and Podgorelec, 2016; Omari and Figueiras-Vidal, 2015; Sesmero *et al.*, 2015], which focus primarily on the construction of complementary classifiers or adaptive weighed combination. These methods generally require all constituent classifiers to be aware of all classes, and thus cannot be directly applied for integrating specialized classifiers.

*Continuous Time Markov Chain (CTMC) based ensemble*, which we propose in this paper, makes no distinction between *ensemble* or *modular* methods. It is a very flexible framework that can *coherently* incorporate both *general classifiers*, *i.e.* those covering the all classes, and *specialized classifiers*, *i.e.* those working with only a subset of classes, into a unified formulation. Here, classifiers can overlap, and each class can be covered by a different number of classifiers. Compared to existing methods, it has two key advantages: (1) It takes into account the relations among predictions when integrating them as well as the different coverage of specialized classifiers. In this way, it can produce more reliable predictions when the ensemble is diverse or unbalanced. (2) It does not rely on any assumptions of the relations among classifiers (*e.g.* there is no requirements that they cover disjoint groups of classes or they form a hierarchy, etc.). This flexibility makes it applicable to a wide range of applications and real-world settings.

It is worth noting that use of stochastic processes has been explored in different machine learning areas. For example, some semi-supervised learning methods [Jaakkola and Szummer, 2002; Xu *et al.*, 2006] rely on Markov random walk for label propagation. Their motivation and task, however, are fundamentally different from our work.

## 3 Classifier Integration Based on CTMC

We consider the task of combining multiple classifiers to make predictions. Suppose we have $M$ classes denoted by $c_1, \ldots, c_M$, and $K$ classifiers, denoted by $f_1, \ldots, f_K$. Each classifier may cover all $M$ classes or just a small subset thereof. Particularly, the subset of classes that $f_k$ works with is given by $\{c_i\}_{i \in I_k}$, where $I_k$ is a set of indexes. When $|I_k| = M$, *i.e.* $f_k$ covers all classes, $f_k$ is called a *general classifier*, otherwise, it is called a *specialized classifier*.

Given a sample $x$, $f_k$ yields an $|I_k|$-dimensional *prediction vector* $\mathbf{p}_k \in \mathbb{R}^{|I_k|}$, in which the entry $\mathbf{p}_k(i)$ with $i \in I_k$ is the probability of sample $x$ belonging to class $c_i$ from the view of classifier $f_k$. Various ways can be used to turn the output of a classifier into such a prediction vector. In the simplest case, where a classifier only yields a class label, $\mathbf{p}_k$ can be constructed as the indicator vector of the predicted class. If a classifier produces scores, *e.g.* SVM, one can turn the scores into a probability vector via a softmax transform. A variety of other classifiers, *e.g.* deep networks with softmax output, can directly yield the prediction vectors by design.

Challenges arise when combining outputs from *specialized classifiers*. First, a specialized classifier, when presented with a sample out of its scope, yields misleading predictions. For instance, a classifier for flower recognition would consider an apple as a certain type of flower. Second, when the coverage of classifiers is unbalanced, *i.e.* some classes are covered by more classifiers than others, the overall prediction may favor classes that are heavily covered. Figure 2 shows an example where the majority voting yields a wrong prediction when individual classifiers predict correctly.

### 3.1 Continuous-Time Markov Chains

We develop a novel methodology to effectively combine specialized classifiers based on a *Continuous-Time Markov*

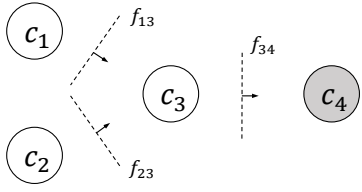

Figure 2: In this example, we have four classes $\{c_1, c_2, c_3, c_4\}$ on a plane, and three classifiers $\{f_{13}, f_{23}, f_{34}\}$, each covering a pair of classes. Decision boundaries are indicated by dashed lines. Given a sample from class $c_4$, both $f_{13}$ and $f_{23}$ would classify it to $c_3$, while $f_{34}$ classify it to $c_4$. Overall, class $c_3$ gets two votes, while the ground-truth $c_4$ gets only one. Hence, simply combining these votes would result in an erroneous prediction $c_3$. To combine specialized classifiers in a more reliable way, we need a new scheme that can take into account the relative preferences among classes.

*Chain (CTMC)* over all classes. In this section, we review some basic properties of CTMC. Formally, a homogeneous CTMC is a stochastic process characterized by a *transition rate matrix* $\mathbf{Q}$, where $q_{ij} = \mathbf{Q}(i, j)$ is the transition rate from state $i$ to $j$ when $i \neq j$, and $-q_{ii}$ is the parameter associated with the exponentially distributed sojourn time in state $i$.

The dynamic of the process can be understood in terms of alarm clocks. When the chain enters state $i$, independent alarm clocks are placed at all other states. The clock in state $j$ will ring after an exponentially distributed time interval with parameter $q_{ij}$. When the first clock rings, the chain goes to that state, discards all alarm clocks and repeats. Hence, it is more likely to transit to a state $j$ with larger $q_{ij}$ value, as the average waiting time $1/q_{ij}$ is shorter. Since the sojourn time in state $i$ is the minimum of a set of exponentially distributed random variables with parameters $\{q_{ij} : j \neq i\}$, it must be an exponential random variable with parameter $\sum_{j \neq i} q_{ij}$. Thus,

$$q_{ii} = -\sum_{j \neq i} q_{ij}. \quad (1)$$

So, to construct a CTMC, we only need to specify the set of transition rates between different states.

When the chain is *ergodic*, $P(X_t)$ will converge to an *equilibrium distribution* $\boldsymbol{\pi}$, which, from an intuitive standpoint, reflects which states it is likely to reach in the long run. $\boldsymbol{\pi}$ does not depend on the initial distribution, and it is related to the transition rate matrix $\mathbf{Q}$ by the *stationary condition*:

$$\boldsymbol{\pi}^T \mathbf{Q} = \mathbf{0}. \quad (2)$$

Equation (2) together with the *normalization constraint*

$$\mathbf{1}^T \boldsymbol{\pi} = 1, \quad (3)$$

where $\mathbf{1}$ is an all-one vector, uniquely determines the stationary probability vector $\boldsymbol{\pi}$. Note that once the $\mathbf{Q}$ is constructed, $\boldsymbol{\pi}$ can be efficiently solved by most linear algebra libraries.

### 3.2 CTMC for Integrating Predictions

The basic idea to integrate specialized predictions is to consider the procedure of reaching the overall prediction as a stochastic process, where one can *walk* from one class to another, following the *preferences* conveyed by individual predictions. One naive implementation of this idea is by tracing.

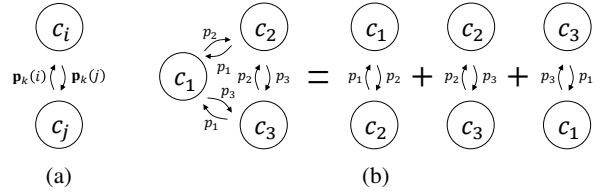

Figure 3: (a) Model the preference between $c_i$ and $c_j$ as a CTMC with two classes as states. (b) The chain for one classifier is the combination of chains representing pairwise preferences.

Specifically, given a sample $x$, the classifier $f_k$ assigns it to a class $c_i$. From this prediction, one can derive $|I_k| - 1$ preferences, given by $\{c_j \to c_i : \forall j \in I_k \setminus \{i\}\}$, meaning that the classifier prefers $c_i$ to all other classes in its domain $I_k$. Take the example in Figure 2 for instance, the classifiers $f_{13}$, $f_{23}$ and $f_{34}$ together result in three preferences: $c_1 \to c_3$, $c_2 \to c_3$, and $c_3 \to c_4$. Following these preferences, one would eventually reach the ground-truth class $c_4$ and stay there, regardless of which class it begins with.

Ideally, when the preferences derived from all classifiers are consistent with each other, it is not difficult to reach a overall prediction by tracing the preferences. However, real-world situations are often more complicated, and individual classifiers may produce contradicting predictions. Besides, this tracing algorithm also discards the rich information in the prediction vector, *i.e.* those relative preferences between *any* pair of classes in the domain. To effectively handle such issues, a more sophisticated scheme is needed. Our approach is to generalize this *deterministic* tracing algorithm into a *CTMC*, where one may walk back and forth between classes guided by the preferences, and we rely on the *equilibrium distribution* to make the final predictions.

We model the pairwise preferences by CTMCs with two states. Given a sample $x$, denote the prediction vector from classifier $f_k$ by $\mathbf{p}_k$. For class $c_i$ and $c_j$ where $i, j \in I_k$, $i \neq j$, we build a chain using the prediction values $\mathbf{p}_k(i)$ and $\mathbf{p}_k(j)$ as transition rates, which is shown in Figure 3 (a). Exploiting Equation (1), the transition rate matrix $\mathbf{Q}$ can be written as

$$\mathbf{Q} = \begin{bmatrix} -\mathbf{p}_k(j) & \mathbf{p}_k(j) \\ \mathbf{p}_k(i) & -\mathbf{p}_k(i) \end{bmatrix}. \quad (4)$$

We can see that $(\mathbf{p}_k(i), \mathbf{p}_k(j))$ is a solution to the stationary condition (2). Thus, the stationary distribution for this chain reflects the relative preference of $f_k$ on these two classes.

Combining all the $\binom{|I_k|}{2}$ preferences into one chain, we can construct a CTMC for classifier $f_k$. The state space consists of all the classes that it covers, and the transition rate from class $c_i$ to $c_j$ is set to the predicted probability for $c_j$. One example with three classes $\{c_1, c_2, c_3\}$ and $\mathbf{p}_k = (p_1, p_2, p_3)$ is illustrated in Figure 3 (b). It is easy to see that the transition rate matrix can be written concisely as

$$\mathbf{Q} = \mathbf{1}\mathbf{p}_k^T - I, \quad (5)$$

where $I$ is the identity matrix. Note that the prediction vector $\mathbf{p}_k$ itself is the solution to the stationary condition (2), as

$$\mathbf{p}_k^T \mathbf{Q} = \mathbf{p}_k^T \mathbf{1} \mathbf{p}_k^T - \mathbf{p}_k^T = 1 \cdot \mathbf{p}_k^T - \mathbf{p}_k^T = \mathbf{0}. \quad (6)$$

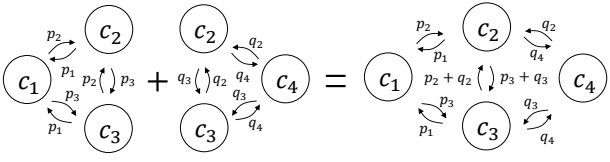

Figure 4: Example of combining two specialized classifiers. Each classifier covers three classes, and they share two classes. For a sample, suppose the output probabilities are $(p_1, p_2, p_3)$ and $(q_2, q_3, q_4)$, respectively. The corresponding CTMC is shown on the left side of the figure. The resulting CTMC is the superposition of the previous two. Note that the transition rates between classes $c_2$ and $c_3$ are the sum of the previous values.

This consistency indicates that merging small chains into a large one is a reasonable strategy for combining preferences.

Next, we combine all the chains from different classifiers by superposition, resulting in a chain over all classes, *i.e.*, the union of individual state spaces. In the resultant chain, the transition rates from $c_i$ to $c_j$ are the *sum* of all the rates defined for the pair. This process is illustrated in Figure 4. The reason for using the sum operator can be seen from the alarm clock analogy. The channel from state $i$ to state $j$ is set with multiple alarm clocks, and the chain transits from state $i$ to $j$ if any of the clocks rings. This is equivalent to set only one alarm clock who will ring after an exponentially distributed time with parameter being the sum of parameters for individual clocks. Overall, the transition rate matrix $\mathbf{Q}$ for the final CTMC can be written as:

$$\mathbf{Q}(i,j) = q_{ij} \triangleq \begin{cases} \sum_{k \in \mathcal{S}_{ij}} \mathbf{p}_k(j) & (i \neq j), \\ -\sum_{j' \neq i} q_{ij'} & (i = j). \end{cases} \quad (7)$$

Here, $\mathcal{S}_{ij} = \{k : i \in I_k \text{ and } j \in I_k\}$ refers to the index set of classifiers that cover both $c_i$ and $c_j$. Given a sample, one can construct $\mathbf{Q}$ and solve the equilibrium distribution $\boldsymbol{\pi}$ as the *final prediction*. In this prediction, the most likely class label is $\hat{i} = \arg\max_i \boldsymbol{\pi}(i)$. Let's revisit the example in Figure 2. For this case, the transition rate matrix is given by

$$\mathbf{Q} = \begin{bmatrix} -1 & 0 & 1 & 0 \\ 0 & -1 & 1 & 0 \\ 0 & 0 & -1 & 1 \\ 0 & 0 & 0 & 0 \end{bmatrix}. \quad (8)$$

Then the prediction is $(0, 0, 0, 1)$, as expected.

### 3.3 Properties of CTMC Formulation

First, we show that the CTMC method is applicable to a reasonable configuration of specialized classifiers. We say a set of classifiers is *connected*, if the graph representation of the constructed CTMC described in Section 3.2 is connected.

**Proposition 1.** *If the classifiers in the ensemble is connected, covers all the classes when combined, and the probabilistic predictions are positive (i.e. predicting non-zero probabilities on all classes that they cover). Then the constructed CTMC is ergodic, and thus CTMC ensemble is applicable and will output a unique prediction.*

*Proof.* The number of states is finite. To show the constructed CTMC is ergodic, it suffices to show that it is irreducible. From the construction of CTMC for each classifier (Figure 3), we know that classes belong to one classifier communicates with each other. Hence, if the classifiers are connected, then all the classes communicate with each other, and thus the irreducibility follows. □

This is a mild condition. For example, it is satisfied for all those classifiers relying on a softmax transform to produce a probabilistic output. The following proposition characterizes the behavior of the CTMC ensemble.

**Proposition 2.** *Given a transition rate matrix $\mathbf{Q}$ defined by Equation (7), the overall prediction $\boldsymbol{\pi}$ has*

$$\boldsymbol{\pi}(i) = \frac{1}{|\mathcal{F}_i|} \sum_{k \in \mathcal{F}_i} \omega_k \mathbf{p}_k(i), \text{ with } \omega_k = \sum_{j \in I_k} \boldsymbol{\pi}(j). \quad (9)$$

*Here, $\mathcal{F}_i$ refers to the set of all classifiers that cover $c_i$.*

*Proof.* Let $a_{ki} = \mathbb{I}(i \in I_k)$, then $q_{ij}$ can be rewritten as

$$q_{ij} = \begin{cases} \sum_k a_{ki} a_{kj} \mathbf{p}_k(j) & (j \neq i) \\ \sum_k a_{ki} a_{kj} (\mathbf{p}_k(j) - 1) & (j = i) \end{cases} \quad (10)$$

Here, we utilize a fact $a_{ki} a_{kj} = a_{ki}$ when $i = j$. Hence, $\boldsymbol{\pi}^T \mathbf{Q}$ can be expanded into

$$(\boldsymbol{\pi}^T \mathbf{Q})(j) = \sum_i \sum_k a_{ki} a_{kj} \mathbf{p}_k(j) \boldsymbol{\pi}(i) - \sum_k a_{kj} \boldsymbol{\pi}(j). \quad (11)$$

With $\boldsymbol{\pi}^T \mathbf{Q} = 0$, we then obtain

$$\boldsymbol{\pi}(j) = \left( \sum_k a_{kj} \right)^{-1} \sum_i \sum_k a_{ki} a_{kj} \mathbf{p}_k(j) \boldsymbol{\pi}(i)$$

$$= \frac{1}{|\mathcal{F}_j|} \sum_{k \in \mathcal{F}_j} \left( \sum_{i \in I_k} \boldsymbol{\pi}(i) \right) \mathbf{p}_k(j). \quad (12)$$

Swapping $i$ and $j$ results in Equation (9). □

Proposition 2 shows that $\boldsymbol{\pi}(i)$ is a weighted combination of $\mathbf{p}_k(i)$, the corresponding entry in the predictions from the classifiers that cover $c_i$. Here, the combination weight $\omega_k = \sum_{j \in I_k} \boldsymbol{\pi}(j)$, which equals the total probability mass that falls in its domain $I_k$, can be interpreted as the *relevance* of the classifier $f_k$ to the given sample. Note that $\boldsymbol{\pi}$ and $\omega_k$ are mutually dependent. The optimal $\boldsymbol{\pi}$ can be considered as the fixed point of Equation (9), where the classifier relevances agree with the overall prediction. Under the special setting when all classifiers cover the same set of classes, we have:

**Corollary 1.** *If every classifier covers the entire class space, then $\boldsymbol{\pi}$ equals the average of individual predictions,*

$$\boldsymbol{\pi} = \frac{1}{K} \sum_{k=1}^{K} \mathbf{p}_k. \quad (13)$$

This corollary follows immediately from Proposition 2. In particular, $\omega_k = 1$ for each $k$ in this case. The next corollary demonstrates the power in stitching specialized classifiers.

**Corollary 2.** *In Proposition 1, if all specialized classifiers are obtained by first restricting results from one given general classifier into their domain and then rescaling properly, then the output of CTMC exactly recovers this general classifier.*

*Proof.* From Proposition 2, we know that the prediction from the general classifier is a solution. This is because in Equation 9, by the construction of specialized classifiers, we have $\pi(i) = \omega_k \mathbf{p}_k(i)$ for all $i$ and $k$. Then by Proposition 1, it is the unique solution. □

For example in Figure 4, suppose that the two prediction vectors are $(p_1, p_2, p_3) = (0.7, 0.1, 0.2)$ and $(q_2, q_3, q_4) = (0.2, 0.4, 0.4)$. On their shared classes $c_2$ and $c_3$, the relative preferences are consistent, *i.e.* $p_2 : p_3 = q_2 : q_3 = 1 : 2$. The output of CTMC is $(7, 1, 2, 2)/12$. We can verify that the two predictions are the restricted version of the output. This shows that CTMC ensemble can find a coherent global picture if all local information collected from specialized classifiers do not contradict with each other.

## 4 Experiment

We tested the proposed method in comparison with mainstream baselines on two commonly used benchmarks, *ILSVRC 2012* [Russakovsky *et al.*, 2015], and *CIFAR-100* [Krizhevsky and Hinton, 2009]. The ILSVRC benchmark is constructed based on ImageNet, which contains $1.2M$ training images in 1000 classes, and $50K$ validation images for testing. The CIFAR-100 dataset consists of $60K$ images in 100 classes, of which $50K$ are for training and the other $10K$ are for testing.

The base classifiers used in our experiments are all deep Convolutional Neural Networks (CNNs), which all have a softmax layer on top of the final representations, and thus can directly output the probabilistic prediction vectors. Particularly, three network architectures are used in our experiments, namely ResNet-56 [He *et al.*, 2016a], pre-activation ResNet-101 [He *et al.*, 2016b], and Inception-ResNet-v2 [Szegedy *et al.*, 2016]. In this way, we can see how well the proposed method can work with different models.

We compared with several commonly used methods for classifier combination: (1) **vote**: each classifier votes for the class with highest probability, and choose the smaller index if there is a tie; (2) **mean**: averaging the prediction scores from *all* individual classifiers as output; (3) **m-mean**: multiplicity-aware mean, where the prediction scores on individual classes are averages over only those covering classifiers; (4) **prod**: geometric mean of the prediction scores from *all* individual classifiers; (5) **m-prod**: multiplicity-aware product, where the geometric mean of prediction scores for each class is taken over only the covering classifiers. Here, *m-mean* and *m-prod* are natural way to mitigate the effect of unbalanced classifier coverage. Following the common practice, we measure the performances by top-1 and top-5 error rates.

### 4.1 Combining Specialized Classifiers

The primary goal of this work is to develop an effective approach to combining *specialized classifiers*, *i.e.* those that cover subsets of classes. The first experiment is to assess the effectiveness of the CTMC method on this task, in comparison with the baseline methods.

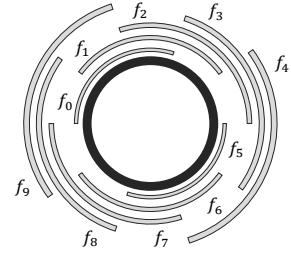

Figure 5: The coverage of specialized classifiers on ILSVRC. Here, the inner circle represents the 1000 classes and the outer arcs indicate the coverage of individual classifiers. We can see that adjacent classifiers considerably overlap with each other.

**On ILSVRC** We first conducted the experiment on ILSVRC. Specifically, for ILSVRC, we train 10 specialized classifiers $f_0, \ldots, f_9$, each covering 300 classes. Adjacent classifiers, *i.e.* $f_k$ and $f_{k+1}$ have 200 classes in common, as shown in Figure 5. For this experiment, we used a trimmed down version of *Inception-ResNet-v2* [Szegedy *et al.*, 2016][1]. We followed the standard practice to train the networks, using random-crop augmentation, RMSProp [Tieleman and Hinton, 2012] with decay 0.9 and $\epsilon = 1.0$, weight decay $0.0005$, and batch size $512$. The learning rate was initialized to $0.45$ and scaled down by a factor $0.1$ per $120K$ iterations. All experiments were terminated at iteration $250K$ to avoid overfitting.

Results obtained by combining different subsets of these specialized classifiers are shown in Table 1. For the first two rows of the table, the classifiers in the subset $\{f_0, f_2, \ldots, f_8\}$ or $\{f_1, f_3, \ldots, f_9\}$ are only minimally overlapped to form a connected graph, which establishes the basic performance. For the rows below, we gradually incorporate additional classifiers to increase the overlap.

From the results, we observed: (1) *m-mean* and *m-prod*, takes into account the factor of multiplicity under the condition of unbalanced coverage, yields better results than naive *vote*, *mean* and *prod*. Nevertheless, the CTMC method consistently outperforms all baseline methods, by considerable margins *w.r.t.* relative reduction of error rates compared to best baselines. (2) When increasing the number of classifiers and particularly the overlaps among them, we can observe steady improvement of the performance. (3) By combining all 10 specialized classifiers with the CTMC method, we obtained a top-1 error at $22.480\%$ and top-5 error at $6.676\%$. To put the results into perspective, we also trained a general classifier using the same network structure and hyperparameters, which yields a top-1 error $24.542\%$ and a top-5 error $7.224\%$. The relative reduction of top-1 and top-5 error rates are respectively $8.4\%$ and $8.2\%$. This is a notable performance gain, considering the baseline was already very strong. This also suggests that combining specialized classifier is a viable way to boost performance.

---
[1]We used a trimmed down version, so it fits our computational budget to train 10 networks. This version contains $3, 6, 3$ blocks in A, B, and C regions as compared to original's $5, 10, 5$.

Table 1: Combining subsets of specialized classifiers $\{f_0, f_1, \ldots, f_9\}$ on the ILSVRC dataset. Classifiers use a trimmed version Inception-ResNet-v2 network structure. Each classifier covers 300 classes, and adjacent classifiers have 200 common classes (Figure 5).

| Classifier Set | Top-1 Error (%) | | | | | | Top-5 Error (%) | | | | | |
|---|---|---|---|---|---|---|---|---|---|---|---|---|
| | vote | mean | m-mean | prod | m-prod | CTMC | vote | mean | m-mean | prod | m-prod | CTMC |
| $\{f_0, f_2, f_4, f_6, f_8\}$ | 61.298 | 37.042 | 28.286 | 29.592 | 28.350 | **25.322** | 15.962 | 12.332 | 11.734 | 13.376 | 11.144 | **7.952** |
| $\{f_1, f_3, f_5, f_7, f_9\}$ | 57.138 | 34.746 | 27.878 | 29.342 | 27.898 | **24.992** | 16.052 | 12.742 | 12.078 | 12.600 | 10.686 | **7.718** |
| $\{f_0, f_2, f_4, f_6, f_8\} \cup \{f_1\}$ | 51.060 | 33.664 | 27.504 | 29.294 | 27.526 | **24.518** | 19.706 | 12.148 | 11.508 | 12.964 | 10.536 | **7.696** |
| $\{f_0, f_2, f_4, f_6, f_8\} \cup \{f_1, f_3\}$ | 47.436 | 35.752 | 26.574 | 28.670 | 26.622 | **24.156** | 23.898 | 11.806 | 11.046 | 12.924 | 10.050 | **7.528** |
| $\{f_0, f_2, f_4, f_6, f_8\} \cup \{f_1, f_3, f_5\}$ | 43.172 | 34.070 | 25.912 | 27.586 | 25.832 | **23.506** | 22.020 | 11.096 | 10.620 | 12.326 | 9.634 | **7.206** |
| $\{f_0, f_2, f_4, f_6, f_8\} \cup \{f_1, f_3, f_5, f_7\}$ | 37.300 | 28.892 | 25.008 | 25.060 | 24.828 | **22.842** | 14.856 | 10.334 | 10.136 | 9.954 | 9.206 | **6.886** |
| $\{f_0, f_1, f_2, f_3, f_4, f_5, f_6, f_7, f_8, f_9\}$ | 30.498 | 24.530 | 24.530 | 24.258 | 24.258 | **22.480** | 14.172 | 10.014 | 10.014 | 8.548 | 8.548 | **6.676** |

Table 2: Combining subsets of 20 specialized classifiers on the CIFAR-100 dataset. They use ResNet-56 structure. Each classifier covers 30 classes and adjacent classifiers share 25 classes.

| Classifier Set | Top-1 Error (%) | | | | | | Top-5 Error (%) | | | | | |
|---|---|---|---|---|---|---|---|---|---|---|---|---|
| | vote | mean | m-mean | prod | m-prod | CTMC | vote | mean | m-mean | prod | m-prod | CTMC |
| $\{g_0, g_5, g_{10}, g_{15}\}$ | 74.60 | 58.04 | 49.59 | 49.95 | 49.68 | **45.51** | 25.82 | 21.83 | 21.48 | 22.31 | 21.39 | **16.64** |
| $\{g_1, g_6, g_{11}, g_{16}\}$ | 74.73 | 55.87 | 49.46 | 50.04 | 49.55 | **43.94** | 26.05 | 21.99 | 21.76 | 22.27 | 21.24 | **15.91** |
| $\{g_2, g_7, g_{12}, g_{17}\}$ | 74.43 | 54.07 | 50.47 | 51.11 | 50.62 | **46.26** | 26.33 | 22.34 | 22.11 | 22.83 | 21.68 | **17.47** |
| $\{g_3, g_8, g_{13}, g_{18}\}$ | 76.06 | 56.66 | 47.96 | 48.36 | 48.03 | **43.76** | 25.60 | 21.74 | 21.43 | 22.17 | 21.32 | **16.86** |
| $\{g_4, g_9, g_{14}, g_{19}\}$ | 73.67 | 54.73 | 49.82 | 50.25 | 49.91 | **46.97** | 25.51 | 22.19 | 21.97 | 22.87 | 22.13 | **17.75** |
| $\{g_0, g_4, g_8, g_{12}, g_{16}\}$ | 65.42 | 55.93 | 44.67 | 45.56 | 44.74 | **41.01** | 23.56 | 21.16 | 20.31 | 21.31 | 19.05 | **15.29** |
| $\{g_1, g_5, g_9, g_{13}, g_{17}\}$ | 64.94 | 55.53 | 48.49 | 49.43 | 48.49 | **44.73** | 22.53 | 20.98 | 20.80 | 23.75 | 21.39 | **17.09** |
| $\{g_2, g_6, g_{10}, g_{14}, g_{18}\}$ | 65.39 | 53.66 | 46.69 | 48.37 | 47.03 | **43.21** | 22.87 | 21.39 | 21.22 | 24.10 | 21.59 | **16.72** |
| $\{g_3, g_7, g_{11}, g_{15}, g_{19}\}$ | 67.19 | 54.86 | 43.66 | 44.64 | 43.69 | **40.88** | 23.10 | 21.30 | 20.79 | 22.04 | 19.80 | **15.78** |
| $\{g_0, g_3, g_6, g_9, g_{12}, g_{15}, g_{18}\}$ | 54.73 | 41.63 | 40.73 | 41.23 | 41.05 | **39.82** | 21.70 | 19.15 | 18.80 | 18.47 | 18.16 | **15.83** |
| $\{g_1, g_4, g_7, g_{10}, g_{13}, g_{16}, g_{19}\}$ | 51.86 | 41.60 | 40.66 | 41.16 | 41.00 | **38.67** | 21.58 | 19.46 | 19.09 | 19.05 | 18.83 | **15.52** |
| $\{g_2, g_5, g_8, g_{11}, g_{14}, g_{17}\}$ | 56.20 | 48.09 | 44.25 | 45.77 | 44.58 | **40.98** | 23.11 | 20.59 | 20.41 | 22.77 | 20.29 | **16.30** |
| $\{g_0, g_2, g_4, g_6, g_8, g_{10}, g_{12}, g_{14}, g_{16}, g_{18}\}$ | 44.81 | 38.30 | 38.30 | 38.93 | 38.93 | **35.81** | 20.70 | 17.54 | 17.54 | 16.44 | 16.44 | **13.18** |
| $\{g_1, g_3, g_4, g_6, g_9, g_{11}, g_{13}, g_{15}, g_{17}, g_{19}\}$ | 44.37 | 37.83 | 37.83 | 38.31 | 38.31 | **35.94** | 20.10 | 17.63 | 17.63 | 16.74 | 16.74 | **13.89** |
| $\{g_0, g_1, g_2, g_3, g_4, g_5, g_6, g_7, g_8, \ldots, g_{19}\}$ | 37.65 | 35.16 | 35.16 | 35.64 | 35.64 | **32.42** | 16.45 | 14.94 | 14.94 | 14.36 | 14.36 | **10.98** |

Table 3: Combining pre-activation ResNet-101 models on the ILSVRC dataset. $h_0, h_1, h_2$ are three general classifiers. $R, S, U$ are three sets of specialized classifiers defined in Sec. 4.2, namely *random split*, *synset split*, and *unbalanced*, respectively.

| Classifier Set | Top-1 Error (%) | | | | | | Top-5 Error (%) | | | | | |
|---|---|---|---|---|---|---|---|---|---|---|---|---|
| | vote | mean | m-mean | prod | m-prod | CTMC | vote | mean | m-mean | prod | m-prod | CTMC |
| $\{h_0\}$ | 22.066 | 22.066 | 22.066 | 22.066 | 22.066 | 22.066 | 21.998 | 6.146 | 6.146 | 6.146 | 6.146 | 6.146 |
| $\{h_0\} \cup R$ | 22.462 | 22.092 | 22.092 | 21.820 | 21.820 | 21.836 | 16.540 | 8.356 | 8.356 | 6.404 | 6.404 | **5.972** |
| $\{h_0\} \cup S$ | 25.240 | 23.278 | 23.278 | 21.782 | 21.782 | 21.708 | 22.600 | 15.292 | 15.292 | 7.670 | 7.670 | **6.108** |
| $\{h_0\} \cup U$ | 43.382 | 31.564 | 23.974 | 21.792 | 21.488 | 21.308 | 25.912 | 16.532 | 15.196 | 9.948 | 8.794 | **5.802** |
| $\{h_1\}$ | 22.066 | 22.066 | 22.066 | 22.066 | 22.066 | 22.066 | 22.006 | 6.050 | 6.050 | 6.050 | 6.050 | 6.050 |
| $\{h_1\} \cup R$ | 22.526 | 22.176 | 22.176 | 21.980 | 21.980 | 21.918 | 16.630 | 8.304 | 8.304 | 6.290 | 6.290 | **5.966** |
| $\{h_1\} \cup S$ | 25.162 | 23.026 | 23.026 | 21.628 | 21.628 | 21.584 | 22.554 | 15.292 | 15.292 | 7.580 | 7.580 | **5.990** |
| $\{h_1\} \cup U$ | 43.304 | 31.522 | 23.882 | 21.740 | 21.484 | 21.222 | 25.928 | 16.486 | 15.126 | 9.918 | 8.758 | **5.730** |
| $\{h_2\}$ | 21.980 | 21.980 | 21.980 | 21.980 | 21.980 | 21.980 | 21.914 | 6.094 | 6.094 | 6.094 | 6.094 | 6.094 |
| $\{h_2\} \cup R$ | 22.446 | 22.074 | 22.074 | 21.886 | 21.886 | 21.820 | 16.534 | 8.330 | 8.330 | 6.352 | 6.352 | **5.982** |
| $\{h_2\} \cup S$ | 25.210 | 23.126 | 23.126 | 21.648 | 21.648 | 21.542 | 22.552 | 15.290 | 15.290 | 7.552 | 7.552 | **5.942** |
| $\{h_2\} \cup U$ | 43.326 | 31.528 | 23.982 | 21.774 | 21.468 | 21.156 | 25.882 | 16.472 | 15.076 | 9.922 | 8.760 | **5.752** |
| $\{h_0, h_1, h_2\}$ | 21.518 | 21.520 | 21.520 | 21.526 | 21.526 | 21.520 | 18.002 | 5.806 | 5.806 | 5.834 | 5.834 | 5.806 |
| $\{h_0, h_1, h_2\} \cup R$ | 21.464 | 21.480 | 21.480 | 21.490 | 21.490 | 21.448 | 14.022 | 7.398 | 7.398 | 5.808 | 5.808 | **5.706** |
| $\{h_0, h_1, h_2\} \cup S$ | 21.334 | 21.414 | 21.414 | 21.232 | 21.232 | 21.140 | 18.200 | 11.842 | 11.842 | 5.922 | 5.922 | **5.678** |
| $\{h_0, h_1, h_2\} \cup U$ | 22.838 | 22.502 | 21.672 | 21.222 | 21.086 | 21.000 | 17.294 | 13.412 | 12.678 | 6.730 | 6.462 | **5.588** |
| $R \cup S \cup U$ | 48.932 | 34.850 | 26.444 | 27.490 | 26.350 | **22.222** | 25.940 | 12.832 | 11.612 | 12.588 | 10.996 | **6.070** |
| $\{h_0, h_1, h_2\} \cup R \cup S \cup U$ | 21.854 | 21.758 | 21.262 | 21.180 | 21.048 | **20.972** | 14.598 | 9.884 | 9.464 | 6.462 | 6.200 | **5.488** |

**On CIFAR-100** On CIFAR-100, we adopted different settings, as this dataset differs significantly from ImageNet. Specifically, we trained 20 specialized classifiers $g_0, \ldots, g_{19}$, each covering 30 classes, and adjacent classifiers have 25 classes in common. As the image size here is much smaller ($32 \times 32$), it suffices to use ResNet-56 as the base network. We followed the common practice to train the networks: data augmentation with random crop, SGD with momentum 0.9 and batch size 128, learning rates initialized to 0.1 and scaled down by 0.1 per $45K$ iterations, and weight decay 0.0005.

Table 2 shows the top-1 error rates obtained by combining different subsets of classifiers using different methods. We observed similar trends on CIFAR-100: CTMC method outperforms baseline methods consistently and significantly. Also, the performance generally improves as we increase the size of the ensemble and thus the overlap among classes (see the rows from top to bottom). Again, we trained a general classifier, which yields a top-1 error rate at $35.51\%$. The best error rate from an CTMC ensemble of specialized classifier is $32.42\%$, about $8.7\%$ of relative reduction. However, we should also note that this reduction is achieved by combining *many* specialized classifiers. As shown in Table 2, when the minimal set of classifiers are combined, *i.e.* 4 specialized classifiers as in the first table block. the performance is not yet comparable to a well-trained general classifier. Albeit it defeats all other combining methods in the same table.

### 4.2 Ensemble Augmentation

We have seen that CTMC provides an effective means to combine specialized classifiers. In the following experiments, we study (1) whether we can obtain *further* performance gains by incorporating *both* general and specialized classifiers, and (2) if this is the case, what are the more effective strategies to configure a specialized ensemble.

We conducted the experiments on ILSVRC, and considered three configurations to generate specialized classifiers: **(1) Random split.** This baseline strategy randomly evenly splits the 1000 classes into 12 groups, and trains 12 specialized classifiers, each for one group. **(2) Synset split.** Each class in ILSVRC corresponds to a synset, which is a node in the wordnet hierarchy [Miller, 1995]. Therefore, we can exploit this hierarchical structure to group the classes, so that semantically related classes are brought together. Walking down along the hierarchy until reaching the nodes with moderate sub-tree sizes, we obtain 12 groups, which are listed below together with the number of classes contained therein: *structures* (55), *materials* (56), *things* (114), *devices* (122), *homeware* (106), *vehicles* (68), *domestic* (121), *invertebrate* (61), *poultry* (59), *mammals* (97), *mollusc* (60), and *others* (81). **(3) Unbalanced.** Both splits above use disjoint groups, and therefore each class will be covered by two classes (one general and one group-specific). In this split, we consider a setting with *unbalanced* coverage. Particularly, we select two large synset nodes, *instrumentality* (348) and *animals* (398), 20 small synsets, whose sizes are around 30, and 16 tiny synsets, whose sizes are below 10. These new synsets together with the 12 synsets in the *synset split* above result in a collection of 50 groups of classes. These groups may overlap with each other, and their sizes range from 4 to 398. In the end, number of classes covered by 1/2/3/4 classifiers is 130/297/487/86.

In this experiment, we trained 3 general classifiers $h_0, h_1$ and $h_2$, and multiple specialized classifiers, each for a group in the splits introduced above. For all these classifiers, including both the specialized classifiers and the general classifiers, we adopt the same pre-activation ResNet-101 architecture, modulo the difference in the last fully-connected layer, which is class number dependent. One same pre-trained model is used for the initialization. This setting is different from the ILSVRC experiment in Section 4.1 because the specialized classifiers in this experiment are *finetuned* from a well-trained general classifier; while those in the previous experiment are trained *from scratch*. Specifically, the training uses SGD with momentum 0.9 and weight decay 0.0005. Batch size is set to 200 for groups with over 50 classes or 128 for those smaller.

From the results shown in Table 3, we observed: (1) The CTMC method outperforms other combination methods in almost all settings. (2) The CTMC method generally obtains performance improvement when adding specialized methods, no matter whether the coverage is balanced or unbalanced. Other methods, including *m-mean* and *m-prod*, will see a significant performance degradation under the unbalanced condition. This clearly suggests a consistent reliability of the CTMC method. (3) The unbalanced setting which involves specialized classifiers trained on groups of different sizes yields the most significant gains using CTMC. This demonstrates the strength of the CTMC method in integrating very complex classifier configurations. Other methods all fail in such setting, resulting in performances even worse than the baseline. (4) If we remove the general classifier from the ensemble, the error rate sees a sharp rise for baseline methods, while for CTMC, the increase of error is quite moderate.

## 5 Conclusion

We presented a novel method for combining specialized classifiers based on the Continuous-Time Markov Chains (CTMC), which can reason about the relations among preferences conveyed by individual predictions, instead of just simply combining them. This allows it to work with diverse ensembles and the cases with unbalanced coverage.

We compared it with commonly used methods on two public benchmarks, ILSVRC and CIFAR-100. The proposed method consistent outperforms other methods under different settings. Especially when the ensemble is diverse and has unbalanced coverage, other methods failed (performing worse than a general classifier) while the CTMC method continues to bring performance improvement. These results clearly demonstrated the merits of the CTMC method – *competitive performance* and *high reliability* in complicated situations.

## Acknowledgements

This work is partially supported by the Big Data Collaboration Research grant from SenseTime Group (CUHK Agreement No. TS1610626) and the General Research Fund (GRF) of Hong Kong (No. 14236516). We would like to thank Xingcheng Zhang for the artistic drawing of Figure 5.

# References

[Breiman, 1996] Leo Breiman. Bagging predictors. *Machine learning*, 24(2):123–140, 1996.

[Dietterich and Bakiri, 1995] Thomas G. Dietterich and Ghulum Bakiri. Solving multiclass learning problems via error-correcting output codes. *Journal of artificial intelligence research*, pages 263–286, 1995.

[E. Mandler, 1988] J. Schurmann E. Mandler. Combining the classification results of independent classifiers based on the dempster-shafer theory of evidence. *Pattern recognition and artificial intelligence*, pages 381–393, 1988.

[Fabian Caba Heilbron and Niebles, 2015] Bernard Ghanem Fabian Caba Heilbron, Victor Escorcia and Juan Carlos Niebles. Activitynet: A large-scale video benchmark for human activity understanding. In *Proceedings of the IEEE Conference on Computer Vision and Pattern Recognition*, pages 961–970, 2015.

[Freund et al., 1996] Yoav Freund, Robert E Schapire, et al. Experiments with a new boosting algorithm. In *ICML*, volume 96, pages 148–156, 1996.

[He et al., 2016a] Kaiming He, Xiangyu Zhang, Shaoqing Ren, and Jian Sun. Deep residual learning for image recognition. In *The IEEE Conference on Computer Vision and Pattern Recognition (CVPR)*, June 2016.

[He et al., 2016b] Kaiming He, Xiangyu Zhang, Shaoqing Ren, and Jian Sun. Identity mappings in deep residual networks. In *European Conference on Computer Vision*, pages 630–645. Springer, 2016.

[Ho, 1998] Tin Kam Ho. The random subspace method for constructing decision forests. *Pattern Analysis and Machine Intelligence, IEEE Transactions on*, 20(8):832–844, 1998.

[Jaakkola and Szummer, 2002] Martin Szummer Tommi Jaakkola and Martin Szummer. Partially labeled classification with markov random walks. *Advances in neural information processing systems (NIPS)*, 14:945–952, 2002.

[Jacobs et al., 1991] Robert A Jacobs, Michael I Jordan, Steven J Nowlan, and Geoffrey E Hinton. Adaptive mixtures of local experts. *Neural computation*, 3(1):79–87, 1991.

[Jordan and Jacobs, 1994] Michael I Jordan and Robert A Jacobs. Hierarchical mixtures of experts and the em algorithm. *Neural computation*, 6(2):181–214, 1994.

[Karakatič and Podgorelec, 2016] Sašo Karakatič and Vili Podgorelec. Improved classification with allocation method and multiple classifiers. *Information Fusion*, 31:26–42, 2016.

[Krizhevsky and Hinton, 2009] Alex Krizhevsky and Geoffrey Hinton. Learning multiple layers of features from tiny images. 2009.

[Krizhevsky et al., 2012] Alex Krizhevsky, Ilya Sutskever, and Geoffrey E Hinton. Imagenet classification with deep convolutional neural networks. In *Advances in neural information processing systems*, pages 1097–1105, 2012.

[Miller, 1995] George A Miller. Wordnet: a lexical database for english. *Communications of the ACM*, 38(11):39–41, 1995.

[Omari and Figueiras-Vidal, 2015] Adil Omari and Aníbal R Figueiras-Vidal. Post-aggregation of classifier ensembles. *Information Fusion*, 26:96–102, 2015.

[Rogova, 1994] Galina Rogova. Combining the results of several neural network classifiers. *Neural networks*, 7(5):777–781, 1994.

[Russakovsky et al., 2015] Olga Russakovsky, Jia Deng, Hao Su, Jonathan Krause, Sanjeev Satheesh, Sean Ma, Zhiheng Huang, Andrej Karpathy, Aditya Khosla, Michael Bernstein, Alexander C. Berg, and Li Fei-Fei. ImageNet Large Scale Visual Recognition Challenge. *International Journal of Computer Vision (IJCV)*, 115(3):211–252, 2015.

[Sesmero et al., 2015] M Paz Sesmero, Juan M Alonso-Weber, German Gutierrez, Agapito Ledezma, and Araceli Sanchis. An ensemble approach of dual base learners for multi-class classification problems. *Information Fusion*, 24:122–136, 2015.

[Sewell, 2011] Martin Sewell. Ensemble learning. *RN*, 11(02):02, 2011.

[Sharkey, 1999] Amanda J. Sharkey, editor. *Combining Artificial Neural Nets: Ensemble and Modular Multi-Net Systems*. Springer-Verlag New York, Inc., Secaucus, NJ, USA, 1st edition, 1999.

[Simonyan and Zisserman, 2014] Karen Simonyan and Andrew Zisserman. Very deep convolutional networks for large-scale image recognition. *arXiv preprint arXiv:1409.1556*, 2014.

[Szegedy et al., 2015] Christian Szegedy, Wei Liu, Yangqing Jia, Pierre Sermanet, Scott Reed, Dragomir Anguelov, Dumitru Erhan, Vincent Vanhoucke, and Andrew Rabinovich. Going deeper with convolutions. In *Proceedings of the IEEE Conference on Computer Vision and Pattern Recognition*, pages 1–9, 2015.

[Szegedy et al., 2016] Christian Szegedy, Sergey Ioffe, Vincent Vanhoucke, and Alex Alemi. Inception-v4, inception-resnet and the impact of residual connections on learning. *arXiv preprint arXiv:1602.07261*, 2016.

[Tieleman and Hinton, 2012] T. Tieleman and G. Hinton. Lecture 6.5—RmsProp: Divide the gradient by a running average of its recent magnitude. COURSERA: Neural Networks for Machine Learning, 2012.

[Wolpert, 1992] David H Wolpert. Stacked generalization. *Neural networks*, 5(2):241–259, 1992.

[Woźniak et al., 2014] Michał Woźniak, Manuel Graña, and Emilio Corchado. A survey of multiple classifier systems as hybrid systems. *Information Fusion*, 16:3–17, 2014.

[Xu et al., 2006] Yunpeng Xu, Xing Yi, and Changshui Zhang. A random walks method for text classification. In *SDM*, pages 340–347. SIAM, 2006.